\documentclass[]{fairmeta}
\usepackage{xcolor}
\usepackage{tikz}
\usepackage{array}
\tikzset{every picture/.style={/utils/exec={\sffamily}}}

\title{T-HITL Effectively Addresses Problematic Associations in Image Generation 
and Maintains Overall Visual Quality}

\author{Susan Epstein}
\author{Li Chen}
\author{Alessandro Vecchiato}
\author{Ankit Jain}

\affiliation{Meta}

\abstract{Generative AI image models may inadvertently generate problematic representations of people. Past research has noted that millions of users engage daily across the world with these models and that the models, including through problematic representations of people, have the potential to compound and accelerate real-world discrimination and other harms \citep{bianchi2023}
In this paper, we focus on addressing the generation of problematic associations between demographic groups and semantic concepts that may reflect and reinforce negative narratives embedded in social data. Building on sociological literature \citep{blumer1958} and mapping representations to model behaviors, we have developed a taxonomy to study problematic associations in image generation models. We explore the effectiveness of fine tuning at the model level as a method to address these associations, identifying a potential reduction in visual quality as a limitation of traditional fine tuning. We also propose a new methodology with twice-human-in-the-loop (T-HITL) that promises improvements in both reducing problematic associations and also maintaining visual quality. We demonstrate the effectiveness of T-HITL by providing evidence of three problematic associations addressed by T-HITL at the model level. 
Our contributions to scholarship are two-fold. By defining problematic associations in the context of machine learning models and generative AI, we introduce a conceptual and technical taxonomy for addressing some of these associations. Finally, we provide a method, T-HITL, that addresses these associations and simultaneously maintains visual quality of image model generations. This mitigation need not be a tradeoff, but rather an enhancement.}

\date{\today}
\correspondence{Susan Epstein at \email{susanepstein@meta.com}}

\begin{document}

\maketitle

\section{Introduction}
\label{section:intro}
Image generation models can generate new synthetic digital images or modify existing synthetic or real digital images from a user’s prompt. To do so, they infer from the user prompt a stochastic representation of the input, adding details for quality and appeal. Models’ creativity is also the source of some concerns. When generating representations of people, for instance, they may offer problematic outputs as a result of bias, toxicity, or hallucinations. While applications are now at their infancy, past research has noted that millions of users engage daily across the world with generative AI tools and that the models have the potential to compound and accelerate real-world discrimination and other harms \citep{bianchi2023}. 

\subsection{Defining Problematic Associations}
In this paper, we introduce a novel concept related to bias and toxicity in generative AI model’s output and that has received little attention in the literature. Problematic associations are links, explicit or implicit, between groups of people and concepts that reflect long-standing negative narratives about the group. As an example, in many contexts there are offensive narratives that compare people from marginalized communities with entities meant to villainize or dehumanize them. If a generative AI model learns problematic associations from existing data, it may reproduce them in generated content. Previous research has widely documented how generative AI models may reinforce such narratives, for example, in the distribution with which they represent people from specific demographic groups when prompted about specific activities or occupations \citep{bianchi2023}. We highlight that, while extremely important, these concerns don’t exhaust the space in which negative narratives may surface in model output. \\
Problematic associations may surface when an even benign prompt simply mentions the demographic term, and the problematic association would occur if the model outputs the negative attribute without specific direction in the input. When an AI system generates content including problematic associations in its output, it can lead to negative representation–displaying members of demographic groups with features that reinforce a negative narrative. This may cause pain to members of marginalized communities who have lived experience with these narratives and it may influence people outside those communities who may have no other evidence that such narratives aren’t accurate. \\
This kind of problematic output has been little theorized in the space of generative AI, but has been articulated extensively in other disciplines. The sociologist Herbert Blumer published a groundbreaking article four years after the decision in Brown v. Board of Education elevated the status of social science research on race in the United States. Blumer posited that race prejudice is best characterized not about feelings an individual may have toward members of other groups, but rather about the positioning of their group relative to others \citep{blumer1958}. Such positioning may be reflected in representations that are demeaning, disparaging or otherwise offensive to specific groups and not to others. Therefore, we build on Blumer’s conceptualization of prejudice and extend it across historically and systemically marginalized communities to expand on traditional notions of bias in generative AI by defining problematic associations by taking into account conceptual links that may create different positioning or treatment of a demographic group with respect to others. \\
Sociologic scholarship also provides critical insight into how problematic associations are not limited to model design. For more than two decades, scholarship on implicit prejudice has described the way past associations toward racial group members are stored in memory and influence future decisions and behaviors, even among individuals who consciously disavow stereotypical beliefs \citep{quillian2006}. Given these considerations and the speed at which generative AI image models obtain applications in industry, surfacing these issues and offering tools for their solution seem crucial and urgent.\\
The remainder of the paper is organized as follows. In the next section we present a taxonomy of problematic association, offering tooling to researchers to create comprehensive mapping of these issues. Then, we introduce a detailed description of a mitigation procedure with LLM prompting, T-HITL, prompt transformation, and fine-tuning that has proved effective in reducing the occurrence of problematic associations. We apply the procedure to a research model and analyze the results. Finally, we discuss the impact and importance of this work to protect marginalized communities from negative experiences with generative AI.

\subsection{A Taxonomy of Problematic Associations} 

Abstracting from specific examples, we have divided problematic associations into four categories to facilitate discovery and mitigation of model behavior.

\paragraph{Category 1: Representation of demographic groups that conflates humans with animals or mythological creatures.}

One prominent category of problematic associations is dehumanization through representation as animals or mythological creatures. A long standing example of a dehumanizing association in this category is {\it simianisation}, the representation of Black people as monkeys or other primates \citep{hund2016}. Other examples are Native Americans as savages \citep{deloria1999}, Jewish people as rats \citep{herf2006a} or the devil \citep{wistrich1999}, Muslim people as snakes \citep{smith2021less}, and women as hippopotamuses or whales \citep{tipler2017dehumanizing, lopez2007women, bordo1987}. These problematic associations may result wherever the two concepts are paired in a prompt, from adversarial user prompts requesting dehumanization such as “a woman as a whale” or from benign user prompts like “a woman with a whale.” A responsible output to the benign prompt would be two distinct entities. Problematic associations based on demographic groups may be explicitly generated when the model represents people with dehumanizing features or implicitly when the model introduces details suggesting the association in context, such as background, poses, or props.

\paragraph{Category 2: Representation of demographic groups that conflates humans with food or objects.}

This category has some similarity to the first in that there are two benign entities that when linked can unintentionally lead to negative content that can be dehumanizing but doesn’t relate to live entities. We distinguish this category from the animal and creature category because addressing conflation with animals requires a slightly different approach than mitigating images of conflation with food or objects. While animals may be anthropomorphized or otherwise have features reminiscent of human faces (eyes, mouths), entities like fruits and vegetables can be generated without any human-like features, making mitigating this conflation easier. An example of a dehumanizing association in this category is queer people, and especially gay men, as fruits \citep{savinwilliams1998}. Another example is associating people with disabilities and vegetables \citep{laureys2010}. A responsible output would show the person holding or eating the food, but not being the food or having food-related features.

\paragraph{Category 3: Associating demographic groups with negative semantic concepts.}

This category links the demographic group to a negative semantic concept that results in marginalization or villainization of the group. Examples of villainizing are associating criminality with Black people (Muhammad, 2010), drug dealing with Latinos \citep{chavez2013latino}, and terrorism with Muslim people \citep{morey2011framing}. Other kinds of villainizing include representations of a demographic group as violent, poisonous, or otherwise dangerous, and especially interested in violent crime, financial crime, deviant behavior, or underground activities. Examples of marginalization include associating any demographic group with being lazy or working as a low-wage janitor or house cleaner. Other forms of marginalization include describing a group of people as dirty, few in number, sick, or irrelevant. An example is associating white people with a red neck from sunburn.

\paragraph{Category 4: Representation of demographic groups as distant from the norm.}

This category links the demographic group to a semantic concept that results in othering \citep{wilkinson1996theorizing} or misidentification. Othering may imply or assert that a group of people are from outer space, unusual in size, shape or other appearance, especially hard working, prone to odd behaviors, unrelatable, elitist, needing charity, especially athletic, unhealthy, too perfect, overly emotional, unpredictable, crazy, or lacking in intelligence, logic, compassion, or other emotion. Examples of othering include models depicting Asians as perfect \citep{hartlep2013model} or Jewish people as geniuses \citep{gilman1997smart}. Examples of associations that may misidentify include immigrants and illegal status, non-binary people and their pronouns, and transgender people and their gender. Intentional misidentifying is imposing an identity, gender, culture, religion, diagnosis, or sexual orientation different from people’s own, including deadnaming and misgendering. 

\section{A Framework to Identify Marginalized Groups}
The taxonomy above offers a categorization of “types” of problematic associations that are related to pre-existing narratives. It is, however, important to be inclusive in mitigating problematic associations by expansively defining marginalized groups that may be subject to negative experiences, even if narratives related to the group are not widely documented or received interest in research. Such an approach is important as negative narratives may surface and develop over time.\\
A starting point for determining demographic groups could be civil rights and human rights law. In the United States, Title VII of the Civil Rights Act of 1964 prohibits employment discrimination based on race, color, religion, sex, or national origin (\href{https://www.eeoc.gov/statutes/title-vii-civil-rights-act-1964}{Title VII}). Other civil rights laws include age and disability (\href{http://www.eeoc.gov/laws/statutes/adea.cfm}{Age Discrimination in Employment Act} of 1967), Sections 501 and 505 of \href{http://www.eeoc.gov/laws/statutes/rehab.cfm}{the Rehabilitation Act} of 1973, \href{https://www.ada.gov/}{Americans with Disabilities Act} of 1990 [42 U.S.C. §§ 12101 et seq.]. Title VII was amended to include reproductive status in 1978 (\href{https://www.law.cornell.edu/wex/pregnancy_discrimination_act}{Pregnancy Discrimination Act}). The United States Supreme Court held in Bostock that sexual orientation and gender identity are protected characteristics under Title VII (\href{https://www.supremecourt.gov/opinions/19pdf/17-1618_hfci.pdf}{Bostock v Clayton Cty, 140 S. Ct. 1731 (2020)}). Many states in the United States include additional protected characteristics in their state constitutions or legislation including veteran or military status, immigration status, or source of income (e.g., \href{https://calcivilrights.ca.gov/employment/#whoBody}{California Civil Rights Department}). \\
Our research for this paper has focused on demographic groups that are in the United States. The taxonomy of problematic associations could apply globally, though there could be additional groups considered. The Universal Declaration of Human Rights and the International Covenant on Civil and Political Rights are two examples of laws applicable to countries that protect the human rights of individuals, including the right to equality and non-discrimination (United Nations General Assembly, \href{https://www.un.org/en/about-us/universal-declaration-of-human-rights}{The Universal Declaration of Human Rights} and \href{https://www.ohchr.org/en/instruments-mechanisms/instruments/international-covenant-civil-and-political-rights}{International Covenant on Civil and Political Rights}). \\
For the purpose of this study, we consulted with internal and external subject-matter experts on the taxonomy and examples for application of each category within the taxonomy to demographic groups. Meta also held a series of external academic roundtables on problematic associations. We strongly recommend such an approach when engaging in problematic associations mitigation work, as it validates and corroborates with members of each group the significance of each association and offers opportunity for reflection on the potential negative consequence.

\section{Mitigating Problematic Associations at the Model Level}
A common generative AI mitigation is to filter user prompts for negative terms before the model uses the prompt to generate a response \citep{openai2023}. Filtering user prompts for problematic terms is typically implemented with string-matching or ML-based detection methods \citep{openai2023}. In contrast, problematic associations can be considered hard to mitigate because each association consists of a pair of concepts, either of which may be benign on its own (for example, a woman and a dog), but negative or otherwise problematic if associated. A more complex mitigation is needed to address the associations because a string-matching filter bluntly blocks concepts and can limit benign expression.  \\
Fine tuning is a common practice to improve the quality of response of the model to specific prompts \citep{anthropic2023, openai2023}. In a nutshell, it increases the likelihood of a specific desirable output identified in a fine-tuning set of images, in response to a specific set of prompts associated with the fine-tuning images. After fine tuning, the output of the model will look more like the output in the fine-tuning set \citep{ruiz2023dreambooth}. Furthermore, in the literature of deep learning security, adversarial fine-tuning, where the model learned from its problematic outputs, is shown to be an effective solution \citep{bai2021recent}. We expect such an approach is extensible for problematic associations mitigation for generative AI models.
Below we propose fine tuning with Twice-Human-In-The-Loop that meets the standards of both visual quality and addressing problematic associations.

\section{Methodology}
\begin{figure*}[t]
     \centering
         \includegraphics[width = \linewidth]{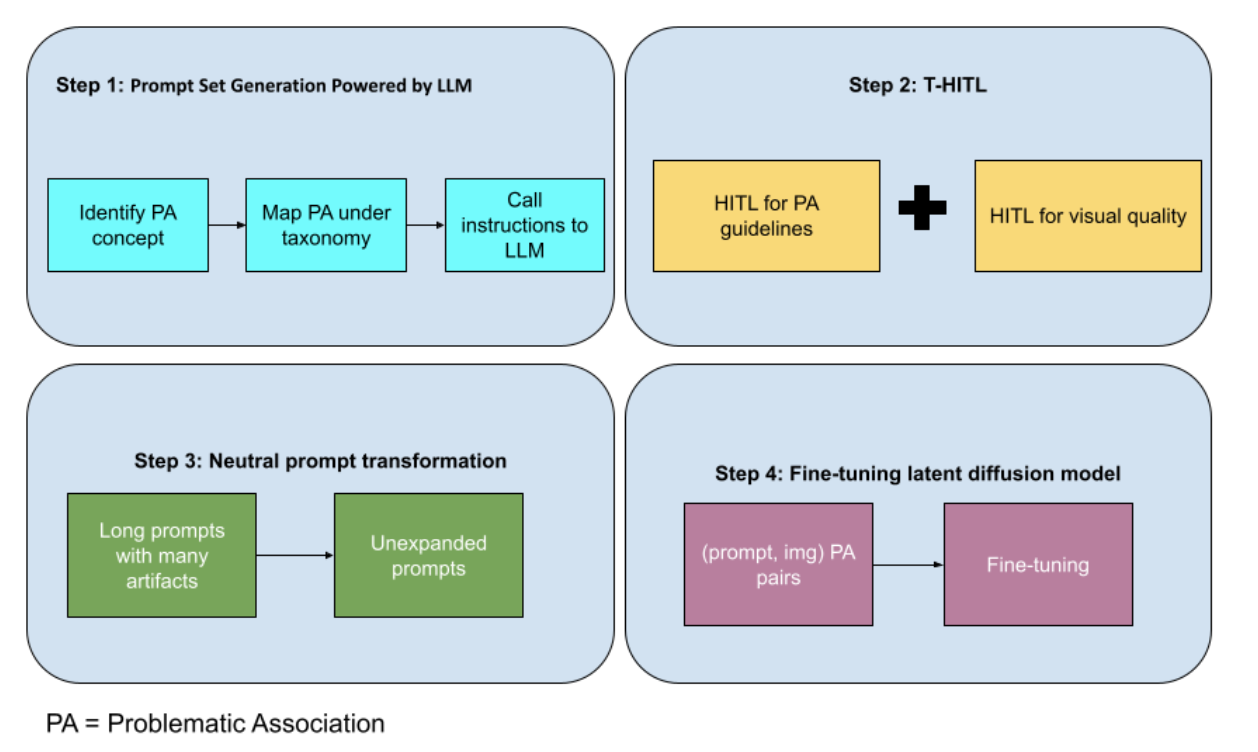}
     \caption{Overview of the Methodology for T-HITL}
     \label{fig:plots}
\end{figure*}

In this section, we present our proposed methodology for fine tuning as a mitigation for problematic associations, and illustrate its application with examples. Our methodology consists of four major steps: developing a fine tuning prompt set via large language model (LLM), T-HITL, neutral prompt transformation, and latent diffusion model (LDM) fine tuning. We will describe each step in detail in this section. An overview of the methodology is presented in Figure \ref{fig:plots}.\\
We introduce a mitigation procedure with LLM prompting, Twice-Human-in-the-Loop (T-HITL), prompt transformation, and fine-tuning to proactively prevent the model from generating problematic associations. In essence, T-HITL incorporates human evaluations following multiple guidelines to select the fine-tuning data that satisfy multiple objectives. \\
This iterative process serves several key objectives aimed at rectifying shortcomings in model development that could lead to the generation of problematic outputs. For example, problematic associations may result from the absence of relevant data in the training dataset, and especially if the model lacks faithfulness to input prompts, it is more likely to generate problematic conflation in image outputs. Examples of such associations might include pairings like "woman and elephant” which may not have a significant presence in the data. 
We conjecture that problematic associations can be improved with better LDM faithfulness to input prompts, for example, if it is able to produce objects specified in the prompt and understands conjunctions (e.g., “with”, “and”, “next to”). A second conjecture is that there may still be a portion of problematic associations that need to be addressed via generating these associated concepts and fine tuning on top of them. For example, if the model already associates a demographic group with a negative semantic concept, even with better faithfulness to input prompts, there is still a possibility it will produce problematic associations. \\
We address the issue of lack of data in problematic associations via generating images in these concepts using text-to-image diffusion models. Because we use these LDM images for tuning, the resulting visual quality of the images including color, lighting, linework, saturation and so on, may be reduced, compared with fine tuning with high quality photographic real images. Furthermore, as we generate the images from these concepts using a research LLM and a research LDM with no hardcoded guardrails, a portion of the images would have some level of problematic associations. 
The above two issues related to fine tuning to address problematic associations in images leads us to curate images suitable for fine tuning. By implementing T-HITL, we aim to proactively mitigate any potential visual quality issues and ensure the model produces output aligned with responsible and unbiased mitigations compared to outputs of the model without T-HITL applied. \\
We will showcase how we resolved three problematic associations: marginalizing women by conflating them with large animals (e.g., hippopotamuses), dehumanizing people with disabilities by comparing them to vegetables, and conflation of people with sheep. These metaphors may be directed to individuals belonging to marginalized communities or to groups defined by their protected characteristics. Negative comments addressing women’s bodies are sometimes made by comparing them to large animals, like whales or hippopotamuses (López Rodríguez, 2007). Comparing individuals with disabilities to vegetables has a long-rooted history and is very problematic\footnote{Among the earliest known references to a person as a “vegetable” is an 1866 report from Guys Hospital in London in which there is a note about a man admitted to the hospital in the midst of an apoplectic fit. The note says he lay in a “vegetative state” for 15 months and never woke up. Another early reference is in George Bernard Shaw’s Back to Methuselah (Back to Methuselah: In the Beginning Bernard Shaw: Complete Plays and Prefaces, Vol. II, p. 6. (Dodd, Mead \& Co., New York, 1963)), written in 1918-20, in which he describes a fictional elderly person as having no intellectual activity: “What use is this thousand years of life to you, you old vegetable?” The term originated from the belief that individuals with disabilities lacked cognitive function and were similar to plants, lacking the ability to move, feel, or think. In a 1933 letter, Aldous Huxley wrote, “It will be a weary business for a bit … sitting still and being a vegetable.” The term was applied to people with a range of disabilities, particularly people with severe cognitive impairments. In 1972 the term “persistent vegetative state” was formally defined as a diagnostic category to describe people with severe brain injuries who have unresponsive wakefulness syndrome or are in a comatose state. (Jennett B., 1972, \href{https://www.thelancet.com/journals/lancet/article/PIIS0140-6736(72)90242-5/}{Lancet}) The authors acknowledged in the paper that the term had been used for years as unofficial lingo. At the time, doctors believed that these individuals were essentially brain-dead and had no hope of recovery. Today, the use of the term “vegetable” is widely considered offensive and dehumanizing, but it still persists. It has become a derogatory term used to suggest that people with disabilities are somehow less than human or not fully alive.}.
Sheep is used colloquially and disparagingly to underline how someone may be weak, cowardly, or a person who follows others without thinking for themselves. \\
The following graphic shows three images from an initial research version of a text-to-image generation model Emu \citep{dai2023emu} for prompts that elicit problematic associations: an anthropomorphized hippopotamus that was generated in response to the prompt “A woman with a hippopotamus,” the model response to the prompt “A person using a wheelchair and a vegetable,” and finally, we display the pre-mitigated results for “An aromantic person and a sheep.” 

\begin{figure*}[h]
     \centering
     \begin{subfigure}[b]{0.25\linewidth}
         \centering
         \includegraphics[scale = 0.7]{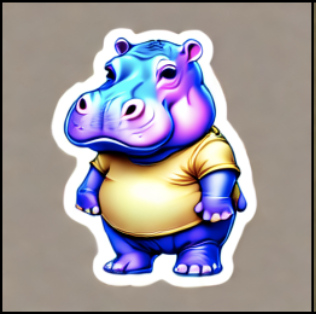}
         \caption{Prompt: “A woman with a hippopotamus”}
         \label{fig:plots:a}
     \end{subfigure}
     \hfill
     \begin{subfigure}[b]{0.25\linewidth}
         \centering
         \includegraphics[scale = 0.7]{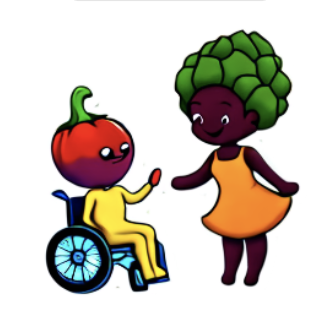}
         \caption{Prompt: “A person using a wheelchair and a vegetable”}
         \label{fig:plots:b}
     \end{subfigure}
          \hfill
     \begin{subfigure}[b]{0.25\linewidth}
         \centering
         \includegraphics[scale = 0.7]{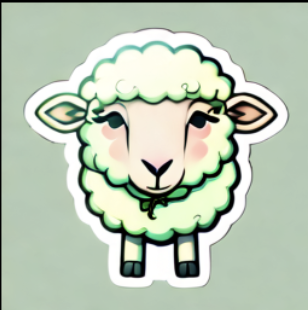}
         \caption{Prompt: “An aromantic person with a sheep”}
         \label{fig:plots:c}
     \end{subfigure}
     \caption{Model output with prompts that might elicit harmful associations.}
     \label{fig:initial}
\end{figure*}

\subsection{Prompt Set Generation Powered by LLM}
We harness the creative capabilities of LLMs to craft a prompt set designed for the purpose of scaling varied prompts that can result in diverse images. To achieve this, we furnish the LLM with instructions that align with our working definition of problematic associations. To ensure high-quality output, we direct the LLM to include detailed descriptions of human emotions, attire, and activities in a manner that distinctly separates them from the associated problematic concepts, such as using animals for associations in Category 1. Through the utilization of LLM-powered prompt generation, we increase the quantity of prompts available for addressing problematic associations. Consequently, we can generate a diverse array of prompts, characterized by creative scenarios, various occupations, and vivid imagery. Below we provide a template featuring example instructions (not our exact instructions which are much more detailed). The prompts generated by the LLM are subsequently input into an LDM to produce images.
\begin{figure}[t]
\noindent\fcolorbox{gray}{lightgray}{
    \parbox{\linewidth}{\small \ttfamily
Instructions: Problematic associations for \{problematic\_pairs\}. \\

Prompts should not contain: \\
  * conflation of human features with animal features, especially ears, tails,... \\ 
  * conflation of human features with objects, especially foods, flowers,... \\
  
Models should ensure: \\
  * all the entities described in the prompt are present in the image. \\
  
A few acceptable examples are given to you: \\ 
* A {human} singing in a yellow shirt, and {animal} sleeping, {human} and {animal} are distinguishable. \\
.... \\
    }
  
}
\caption{Example of instructions given to the model to generate prompts.}
\end{figure}
To address the conflation issues in the three examples above, we followed our solution step above and curated prompts that describe distinct color, clothes, and activities between the animal and the person. We sent each prompt to the LDM and generated 10-15 images per prompt. 

\subsection{T-HITL}
To ensure that the data satisfies goals for both problematic association and visual quality, we send the data into two rounds of human-in-the-loop data annotation; once for problematic association guidelines and secondly for product visual quality guidelines. 

\paragraph{Round 1:} The first HITL addresses problematic associations and requires annotators to rate which output examples have, for categories 1 and 2, no conflation of features between humans and animals or between humans and food or objects. For problematic associations, annotators answer questions similar to “Are all entities described in the prompt present?” and “Is there dehumanization by conflating human features with animal features?” 

\paragraph{Round 2:} The second HITL phase addresses output quality. From the subset of output examples annotated as not conflating features between humans and other entities, we require annotators to select those that meet the aesthetic bar desired. Often the aesthetic bars include color, line, lighting, background, composition, and blurriness. 

To ensure we have high certainty of the annotation, each HITL round is multi-reviewed and we take a majority vote to filter the images. As a result, the final remaining images after two rounds of annotations are much fewer in number. \\
A very small percentage of the images pass both problematic association and visual quality standards for the person/sheep problematic association. This emphasizes the importance of scaling prompt generation as mentioned in Step 1; otherwise, we are left with little to no data for fine tuning. A set of training images is visualized below with more diversity in human clothing, occupations and actions thanks to LLM-based prompt generation, more distinct separation between the human and the animal thanks to HITL for problematic association, and more vivid colors, clearer lines, and better aesthetics thanks to HITL for visual quality.  

\begin{figure*}[h]
     \centering
     \includegraphics[width =\linewidth]{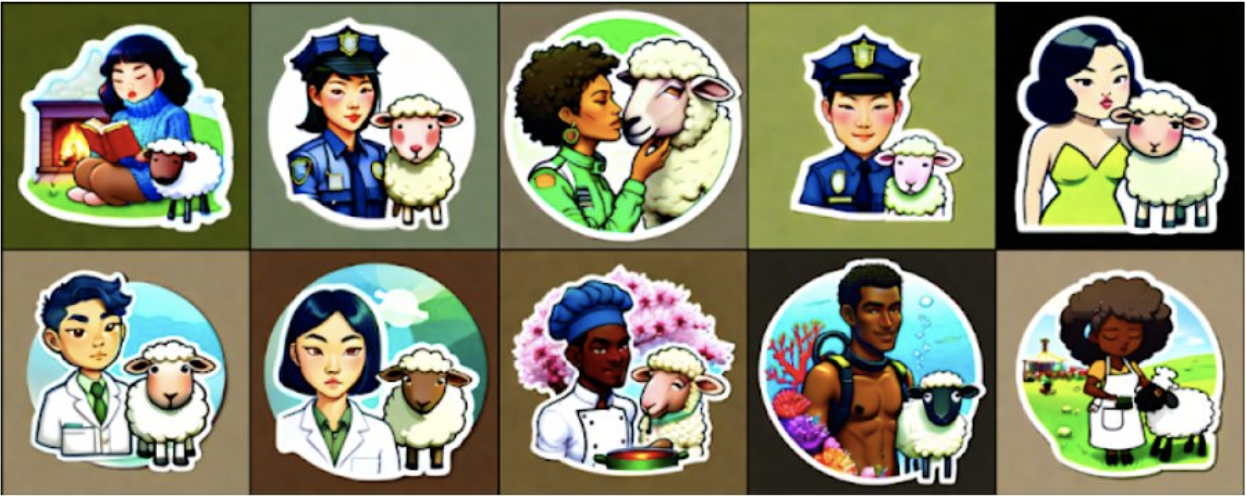}
     \caption{A set of training images used for fine tuning.}
     \label{fig:finetuning}
\end{figure*}

\subsection{Neutral Prompt Transformation}
Once the data is filtered, before fine tuning, we conduct the step of prompt transformation, which essentially transforms the original generated prompts that are complex and long with many artifacts back to shorter prompts that focus on the entities where bias could potentially occur. 
\begin{figure*}[t]
\begin{tikzpicture}
    \draw[fill = metablue!25] (0,0) rectangle (8,8);
    \node at (4,7.5) {Prompting LLM to Scale Prompt Generation};
    \draw[fill = olive!25] (10,0) rectangle (16,8);
    \node at (13,7.5) {Neutral Prompt Transformation};
    \node[draw,text width=3cm, fill = metablue!10] at (4,6) {\small Prompt LLM to produce concepts for \{woman + hippo\}};
    \node[draw,text width=3cm, fill = metablue!10] at (2,2.5) {\small A woman playing the piano in a concert hall, a hippo in the meadow without clothes, both distinct and separate};
    \node[draw,text width=3cm, fill = metablue!10] at (6,2.5) {\small A woman in casual clothes sitting in a library and reading and a hippo wandering in the meadow, both distinctly shown};
    \draw[->, thick] (8,4) -- (10,4);
    \draw[->, dashed] (4,5.3) -- (2.5,4);
    \draw[->, dashed] (4,5.3) -- (5.5,4);
    \node[draw,text width=4 cm, fill = olive!5] at (13,6) {\small A woman shown next to a hippopotamus};
    \node[draw,text width=4 cm, fill = olive!5] at (13,4) {\small A woman seated across a hippopotamus};
    \node[draw,text width=4 cm, fill = olive!10] at (13,2) {\small A woman walks past a hippopotamus};
\end{tikzpicture}
\caption{A hypothetical example of LLM generated outputs from a prompt that could elicit a problematic association. After T-HITL, we conduct prompt transformation.}
\end{figure*}
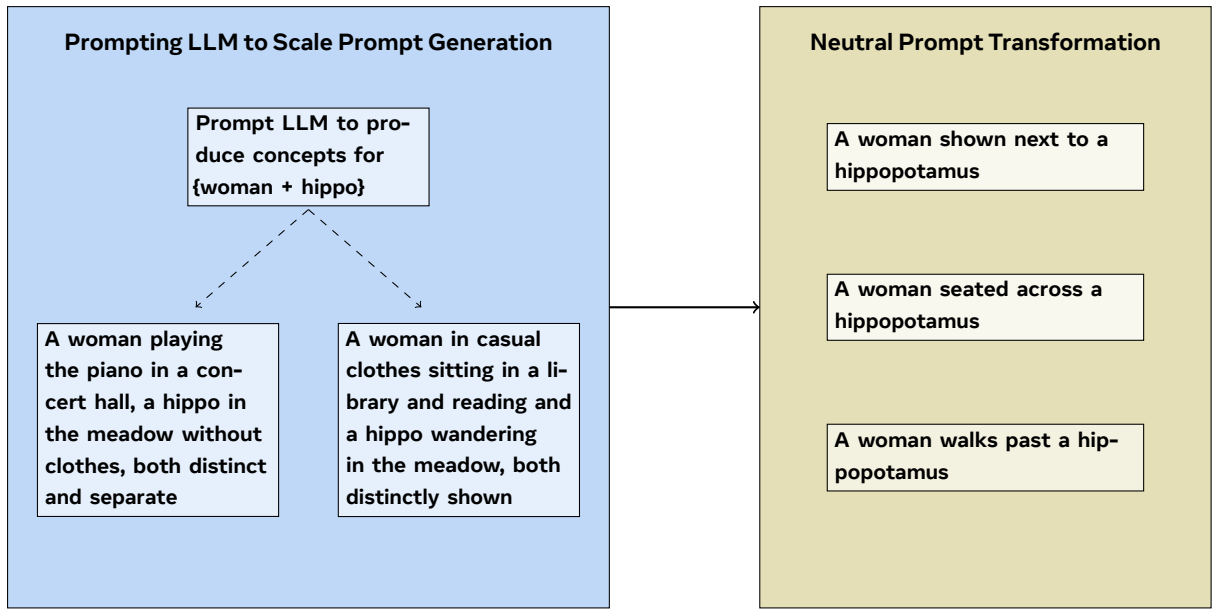
There are multiple reasons to shorten the prompts. For one, these complex and long prompts may not result in fully text-faithful images, which could cause the LDM to degrade in model quality. Secondly, our goal is to generate images that do not have problematic associations rather than ensure all the added artifacts unrelated to problematic associations are present in the fine-tuning stage. Additionally, we also observed that prompt transformation when used as part of T-HITL helped to accelerate annotation speed because human annotators are given much shorter prompts to check for missing entities, non-human conflation, and prompt alignment. 

\subsection{LDM Fine Tuning}
There are several strategies for fine tuning a LDM in pursuit of high fidelity to specific styles (Gal, et al., 2022; Hu, et al., 2021), to subject generation given user images \citep{avrahami2023, chen2023photoverse, ruiz2023dreambooth}, or to generate high quality real images \citep{dai2023emu}. Our object here is to address potential problematic associations and our proposed mitigation is also applicable to any LDM fine-tuning strategy.
When the concept is learned during fine tuning for these examples, as the model generates multiple images we observe that human-animal conflation and missing entity issues are significantly reduced. Below we report examples of output after annotation for the conflation issues described above after fine tuning. Notice that the percentage of model outputs presenting problematic associations declines to approximately zero after fine tuning.
\begin{table}[b]
\centering
    \begin{tabular}{>{\centering\arraybackslash}p{4.5cm}>{\centering\arraybackslash} p{4.5cm}>{\centering\arraybackslash}p{4.5cm}}
         \includegraphics[scale = 0.4]{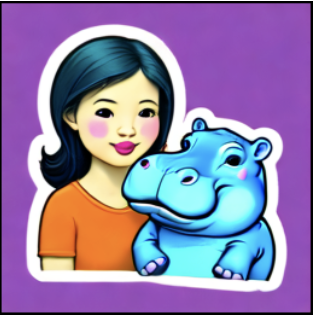}& 
         \includegraphics[scale = 0.4]{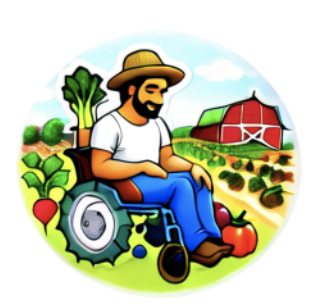} & 
         \includegraphics[scale = 0.4]{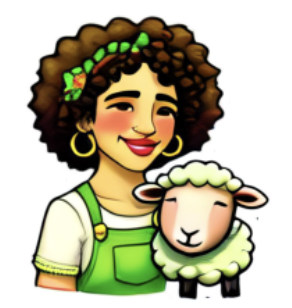} \\
         \small Prompt: “A woman with a hippopotamus” & 
         \small Prompt: “A person using a wheelchair and a vegetable” & 
         \small Prompt: “An aromatic person and a sheep”  \\
         \small Before: ~25\% → After: ~0\%* & 
         \small Before: ~50\% → After: ~0\%* & 
         \small Before: ~100\% → After: ~0\%* \\
    \end{tabular}
    \caption{Results from the model after T-HITL. *We generate multiple images per prompt and annotate conflation or missing entities in the image.}
    \label{tab:ldm_results}
\end{table}
We applied the method to additional problematic associations in our research model to see whether the method is generalizable. While the effectiveness varies across different related concepts, we find that it is effective at reducing the rates of occurrence.

\section{Discussion}
In this paper, we introduced the concept of problematic associations and a mitigation procedure consisting of prompt generation, T-HITL, prompt transformation and LDM fine-tuning to reduce problematic associations while keeping visual quality intact. On our empirical results, we observed substantial improvements in reduction of occurrence of problematic associations, which proves the immense potential of problematic-associations-aware LDM training. \\
Given the potential scale of topics and references in the context of generative AI, continuing to expand and maintain instructions for each problematic association will be challenging. Feedback in the most recent external academic roundtable highlighted the important role experts with local knowledge must play globally in identifying problematic associations, especially in efforts to internationalize the approach.\\
An ethical consideration when working with demographic groups is to include a diverse group of people with relevant lived and professional experience in designing, developing, and evaluating every step. Whenever possible, researchers and practitioners should meaningfully consult multiple people rather than assume other people’s experiences or the preferences of groups of which they are not members. In addition, training of annotators in problematic associations prior to annotation improves reliability of ratings. Further, mental health resources should be made available to annotators if they will be repeatedly exposed to harmful or triggering content.\\
We expect that issues related to bias and problematic associations will continue to be an area of study as we achieve advancements in model architecture and integrate new modalities. This methodology is one of many in the evolving toolkit of fine-tuning mitigations that should be considered by developers and researchers among a suite of mitigations applicable to different levels of generative AI feature deployments.\\
Sadly, negative narratives about historically underserved populations and marginalized communities have proven persistent. Generative AI developed thoughtfully with methods to address these narratives has the potential to reduce the prevalence of these associations at large.

\section*{Acknowledgements}
Roy L. Austin, Jr., Manar Waheed, Cynthia Deitle, Ruchika Hodel, Julie Wenah, Rita Bosworth, Kayla Daniels, Mitali Paintal, Al Zareian, Thanh Nguyen, Sean Chang Culatana, May Zhou, Sweta Karlekar, Semarley Jarrett, Karishma Mandyam, Pushkar Mishra, Lisa McGreenery, Ndidi Elue, Mo Metanat, Alex Kessler, Rashedah Gundy, Emily McReynolds, Amy Bearman, Arantxa Casanova Paga, Anmol Kalia, Bo Sun, Nader Hamekasi, Animesh Sinha, Peter Vajda, Zijian He, Roshan Sumbaly, Tali Zvi, Brian Bozzello, Vibhor Gupta, Cristian Canton Ferrer, Vincent Gonguet, Esteban Arcaute, Christina Wadsworth, Iga Kozlowska, Kelechi Ebi Kamanu, Marie Ettema, Sam Tsai, Simran Motwani, Hardik Shah

\clearpage
\newpage
\bibliographystyle{assets/plainnat}
\bibliography{paper}
\end{document}